\documentclass[A4paper, 11 pt, conference]{ieeeconf}  
\IEEEoverridecommandlockouts       
\usepackage{geometry}
\usepackage{graphics} 
\usepackage{epsfig} 
\usepackage{mathptmx} 
\usepackage{times} 
\usepackage{amsmath} 
\usepackage{amssymb}  
\usepackage{subfigure}
\usepackage[linesnumbered,ruled,vlined]{algorithm2e}
\usepackage{multirow} 

\usepackage{booktabs}

\usepackage{cite}

\usepackage{color}

\usepackage{algpseudocode}

\begin{document}

\title{Autonomous Removal of Perspective Distortion of Elevator Button \\ Images based on Corner Detection}

\author{Nachuan~Ma$^{1}$, Jianbang~Liu$^{2}$, and~Delong~Zhu$^{2}$ 

\thanks{$^{1}$Nachuan Ma is with the College of Electronic and Information Engineering, Tongji University in Shanghai, China, (email: {manachuan@163.com}).
}
\thanks{$^{2}$The authors are with the Department of Electronic Engineering, The Chinese University of Hong Kong, Shatin, N.T., Hong Kong SAR, China, (email: {\{henryliu, zhudelong\}@link.cuhk.edu.hk}).
}

}

\maketitle 
\thispagestyle{empty}

\begin{abstract}
Elevator button recognition is a critical function to realize the autonomous operation of elevators. 
However, challenging image conditions and various image distortions make it difficult to recognize buttons accurately. 
To fill this gap, we propose a novel deep learning-based approach, which aims to autonomously correct perspective distortions of elevator button images based on button corner detection results. 
First, we leverage a novel image segmentation model and the Hough Transform method to obtain button segmentation and button corner detection results.
Then, pixel coordinates of standard button corners are used as reference features to estimate camera motions for correcting perspective distortions. 
Fifteen elevator button images are captured from different angles of view as the dataset. 
The experimental results demonstrate that our proposed approach is capable of estimating camera motions and removing perspective distortions of elevator button images with high accuracy.
\end{abstract}

\section{Introduction}
Autonomous elevator operation is a promising solution for mobile navigation in office buildings. The system consists of three parts: button recognition, motion planning, and robot control. 
Among them, button recognition is the most basic but challenging part. 
Its performance directly determines the success rate and robustness of the entire elevator autonomous operating system.
Traditional button recognition algorithms tend to place markers on the elevator button panel in advance, and then the position of each button relative to the markers can be acquired by calculating the geometric relationship between them.
Unfortunately, these hand-engineered algorithms are inconvenient and will fail if the elevator button panel cannot be marked beforehand.
To overcome this limitation, in the past few years, researchers have proposed plenty of deep learning-based button recognition algorithms \cite{zhu2021ocr,dong2017autonomous,liu2017recognizing}, which can output button recognition results directly from raw elevator button images.
However, the recognition accuracy of these deep learning-based algorithms is not satisfactory due to various image conditions and distortions.
There exist many kinds of button shapes, button sizes, elevator panel designs, and light conditions. 
Meanwhile, various perspective distortions and unexpected blurs make it more challenging to recognize buttons accurately.
In this article, we propose a novel deep learning-based approach to autonomously correct perspective distortions of elevator button images based on button corner detection results.

\begin{figure}
    \flushleft
    \subfigure[original image]{
    \begin{minipage}[t]{0.45\linewidth}
    \centering
    \includegraphics[height=3.7cm,width=3.7cm]{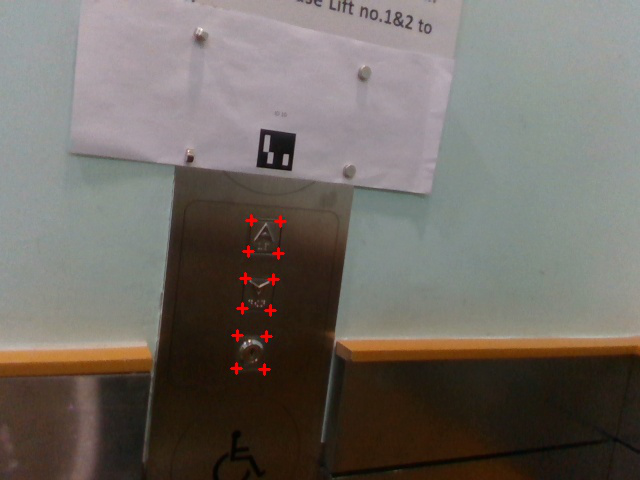}
    \end{minipage}
    }
    \subfigure[corrected image]{
    \begin{minipage}[t]{0.45\linewidth}
    \centering
    \includegraphics[height=3.7cm,width=3.7cm]{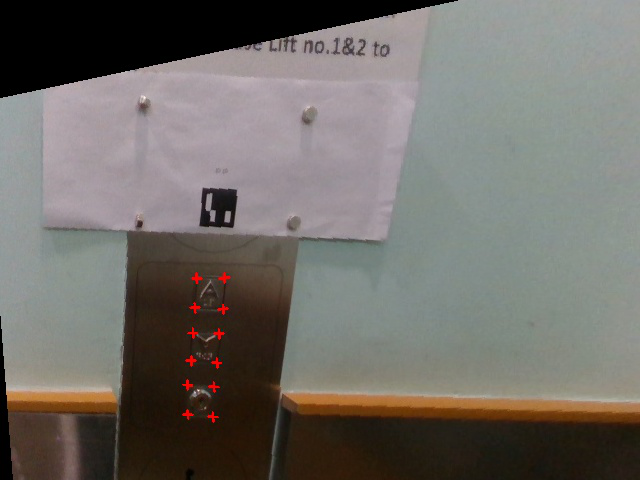}
    \end{minipage}
    }
    \centering
    \caption{A comparison between the original image (left) and the corrected image (right). The red crosses represent the corners of every elevator button.}
    \label{fig1}
\end{figure}

The proposed approach consists of two parts.
The first part is a button corner detection algorithm.
We train an image segmentation model to perform feature extraction of raw elevator button images and obtain button segmentation results.
Then four lines of every button are identified by using the Hough Transform method \cite{duda1972use}.
Pixel coordinates and the order of all button corners can be obtained as they are the intersections of identified lines.
The second part is a pose estimation algorithm.
It takes the hypothetical button corners with standard pixel coordinates as reference points and calculates the camera motions to align corners of raw elevator button images and the reference points.
Then by using an inverse transformation, new elevator button images without perspective distortions can be generated.

The contributions of this work are summarized as follows:

\begin{itemize}
\item We derive detection results of button corners by utilizing an image segmentation model and the Hough Transform method. 
\item We propose a novel algorithm that can autonomously remove perspective distortions of elevator button images based on the detection results of button corners.
\end{itemize}

The remainder of this article is organized as follows. The previous work on elevator button recognition and existing distortion removal methods are reviewed in Sec. II. Sec. III and Sec. IV outline the proposed autonomous perspective distortion removal approach, while the experimental results are presented and discussed in Sec. V. Finally, we draw some conclusions and discuss the future work in Sec. VI.

\section{Related Work}

\subsection{Elevator button recognition}

Before deep learning techniques are widely used in the research area of object recognition for robotics, researchers tend to develop hand-engineered approaches based on traditional image processing techniques to recognize elevator buttons.
For instance, Klingbeil \textit{et al.} \cite{klingbeil2010autonomous} designed a pipeline to realize the function of button detection and character recognition, using a grid fitting method to regress button locations based on a sliding window-based object detector.
This method achieved an accuracy of 86.2\% in a test set containing 50 images.
However, images in the test set were assumed to be in good light conditions without any perspective distortion. 
As a result, the pipeline designed by \cite{klingbeil2010autonomous} cannot be used in natural scenes. 
Zakaria \textit{et al.} \cite{zakaria2014elevator} developed a framework for vision-based elevator external button recognition and localization based on Sobel operator edge detection technique and Wiener filter. 
In \cite{kim2011robust}, the template matching combined with a homography-based transform was used to deal with vision-based button recognition for a robot arm manipulating the elevator.
However, approaches in \cite{zakaria2014elevator}\cite{kim2011robust} were not robust to noise or environmental variability due to the limited capacity of traditional image processing methods.

Various deep learning-based methods have recently been applied to elevator button recognition with the revolution of computational technologies.
The accuracy of elevator button recognition can be significantly improved with the discrimination capabilities of deep neural networks.
For instance, in \cite{islam2017elevator}, the recognition task was formalized as a classification problem, and a hybrid button classification system was proposed, which combined histogram of oriented gradients (HOG), bag-of-words (BoW), and artificial neural networks (ANN). 
Experimental results in \cite{islam2017elevator} showed that ANN could help improve the performance of button classification a lot.
Dong \textit{et al.} \cite{dong2017autonomous} proposed a button recognition system based on convolutional neural networks (CNN), which can achieve a high recognition accuracy for known elevator button panels.
In \cite{liu2017recognizing}, elevator button recognition was regarded as a multi-object detection problem, and a single-shot multi-box detector (SSD) was used as the detection network.
Zhu \textit{et al.} \cite{zhu2021ocr} proposed a novel algorithm for elevator button recognition, called OCR-RCNN, which integrated a character recognition branch into the Faster-RCNN and turned multi-object detection problem into a binary button detection task and a character recognition task. 
Inspired by \cite{liu2021large}, in this article, we design a more advanced semantic segmentation model based on the Deeplabv3+ model \cite{chen2018encoder} and the Hough Transform method to obtain button segmentation and button corner detection results. 


\subsection{Removal of perspective distortions}
In contrast to the vast literature on various elevator button recognition algorithms based on traditional image processing methods or deep learning models, only countable publications studied the removal of perspective distortions for elevator button images.
Researchers have proposed some perspective distortion removal algorithms for document images \cite{takezawa2016camera,liu2015restoring,shafii2015skew}, electroluminescent images \cite{mantel2018correcting}, lithographic watermarked authentication images \cite{xie2014geometric}, and so on.
Zhu \textit{et al.} \cite{zhu2019autonomous} proposed a novel perspective distortion removal algorithm that leveraged the  Gaussian Mixture Model (GMM) and EM framework. 
The algorithm in \cite{zhu2019autonomous} took as input the outcomes of the button center recognizer and finally generated the corrected images.
However, the algorithm in \cite{zhu2019autonomous} can only handle internal panel images that contain a number of buttons and may easily fail for external elevator button images with few button samples.
We further integrate button corners as feature points in this article to realize autonomously perspective distortion removal for robotic elevator button images.
The experimental results demonstrate that our proposed approach can handle external elevator panels well.

\section{Button Corner Detection}
\subsection{Button Segmentation}

\begin{figure}[htpb]
\flushleft
\includegraphics[height=5.3286cm,width=8.3536cm]{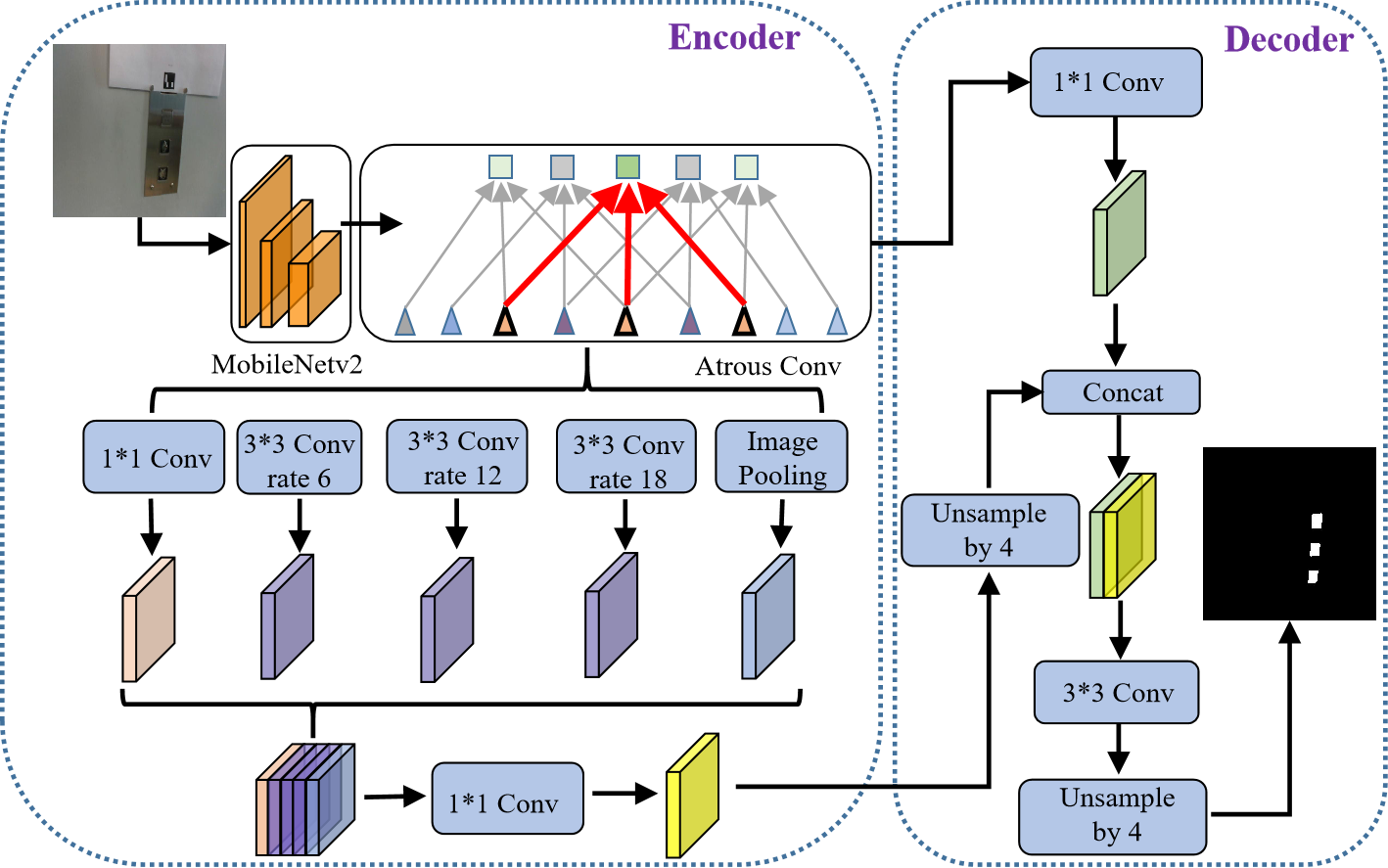}
\caption{The proposed image segmentation model. The input is a raw elevator button image while the output is button segmentation result.}
\label{fig2}
\end{figure}

In this article, we design an image segmentation model based on the Deeplabv3+ model to obtain segmentation results of pixels belonging to elevator buttons. 
The Deeplabv3+ model is one of the state-of-the-art models using advanced deep learning technologies to generate semantic segmentation results of input images, which combines the encoder-decoder structure and the spatial pyramid pooling module.
The encoder-decoder structure can help extract sharp object boundaries, and the spatial pyramid pooling module can help capture rich contextual information. 
The detail of the proposed image segmentation model is shown in Fig. \ref{fig2}.
The input is a raw elevator button image, and the output is a gray-scale image with button segmentation results.

In the encoding stage, different from the deeplabv3+ model, we utilize the MobileNetv2 \cite{sandler2018mobilenetv2}, a depthwise separable backbone, to first extract low-level features and high-level features.
Then several atrous convolutions \cite{chen2017rethinking} with different rates are applied to capture rich semantic information from high-level features.
In the decoding stage, the low-level features module is concatenated with the output of the encoder first.
Then, a $3*3$ convolution module is used to further fuse the extracted features.
Finally, bilinear interpolation is used to obtain the segmentation prediction results of the same size as the input image.

The value of every pixel represents which category it belongs to. For instance, when this button segmentation model is applied to distorted images Fig. \ref{fig1} (a), there exist four categories: `up', `down', `keyhole', and `non-button'.

\subsection{Corner coordinates detection}
After obtaining button segmentation results of raw elevator button images, we first use dilation and erosion methods to reduce image noise and smooth the edges of buttons to improve the performance of line detection. The process of erosion followed by dilation is called a closed operation, which is used to connect neighboring objects and smooth their boundaries at the same time without significantly changing their area.

\begin{figure}[htpb]
\flushleft
\includegraphics[height=3.25cm,width=9cm]{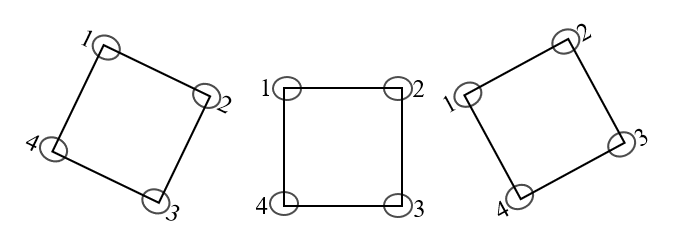}
\caption{The order of corners on a button.}
\label{fig3}
\end{figure}

Then the Hough Transform method is applied to detect four lines of buttons. Hough transform is one of the primary methods to detect geometric shapes from images in computer vision, image analysis, and digital image processing. 
The transformation between two coordinate spaces is to map a curve or line with the same shape from one space to another coordinate space and form a peak. 
Finally, after obtaining detection results of four lines of the button, we can derive pixel coordinates of button corners as they are the intersections of the detected lines. And the order of corners of every button is defined in advance to facilitate the perspective distortion removal algorithm. The order is shown in Fig. \ref{fig3}.

\section{Perspective Distortion Removal}
To begin with, we first define the notations which will be frequently used in this paper. Throughout this work, matrices are written as boldface uppercase letters, and vectors are written as boldface lower letters.
The following notations are used:
\begin{itemize}
\item ${\textbf{C  =  [}}{{\textbf{c}}_1}{\rm{, \cdot \cdot \cdot , }}{{\textbf{c}}_N}{\rm{]  }} \in {R^{2 \times N}}$ - the detected button corners in the image plane;
\item $\mathop {{\textbf{C }}}\limits^ \wedge  {\rm{ =  [}}{\mathop {\textbf{c}}\limits^ \wedge  _1}{\rm{, \cdot \cdot \cdot , }}{\mathop {\textbf{c}}\limits^ \wedge  _N}{\rm{] }} \in {R^{3 \times N}}$ - the detected button corners in the normalized image plane;
\item ${\textbf{U  =  [}}{{\textbf{u}}_1}{\rm{, \cdot \cdot \cdot , }}{{\textbf{u}}_N}{\rm{]  }} \in {R^{2 \times N}}$ - the presupposed standard button corners without distortion in the image plane;
\item $\mathop {\textbf{U}}\limits^ \wedge  {\rm{  =  [}}{\mathop {\textbf{u}}\limits^ \wedge  _1}{\rm{, \cdot \cdot \cdot , }}{\mathop {\textbf{u}}\limits^ \wedge  _N}{\rm{]  }} \in {R^{3 \times N}}$ - the presupposed standard button corners without distortion in the normalized image plane;
\item ${\textbf{G  =  [}}{{\textbf{g}}_1}{\rm{, \cdot \cdot \cdot , }}{{\textbf{g}}_N}{\rm{]  }} \in {R^{2 \times N}}$ - the rectified button corners in the image plane;
\item $\mathop {{\textbf{G }}}\limits^ \wedge  {\rm{ =  [}}{\mathop {\textbf{g}}\limits^ \wedge  _1}{\rm{, \cdot \cdot \cdot , }}{\mathop {\textbf{g}}\limits^ \wedge  _N}{\rm{] }} \in {R^{3 \times N}}$ - the rectified button corners in the normalized image plane;
\item ${{\rm{M}}_{{\mathop{\rm int}} }}\, = \,\left[ \begin{array}{l}
{\rm{F/}}{{\rm{s}}_x}\quad {\rm{0}}\quad \,{{\rm{o}}_x}\\
\;{\rm{0}}\quad \,\,{\rm{F/}}{{\rm{s}}_y}\;\,{{\rm{o}}_y}\\
\;{\rm{0}}\quad \,\;\;\;{\rm{0}}\quad \,{\rm{1}}
\end{array} \right]$ - the intrinsic parameter of the camera;
\item ${\rm{F}}$ - focal length in the meter for fixed focal length, non-zoomed camera;
\item ${{\rm{o}}_x}{\rm{,}}\,{{\rm{o}}_y}$ - image center in pixel;
\item ${{\rm{s}}_x}{\rm{,}}\,\,{{\rm{s}}_y}$ - pixel width and height in meter;
\item ${\textbf{D  =  [}}{{\textbf{d}}_1}{\rm{, \cdot \cdot \cdot , }}{{\textbf{d}}_N}{\rm{]  }} \in {R^{3 \times N}}$ - the spatial coordinates of detected button corners;
\item ${\textbf{E  =  [}}{{\textbf{e}}_1}{\rm{, \cdot \cdot \cdot , }}{{\textbf{e}}_N}{\rm{]  }} \in {R^{3 \times N}}$ - the spatial coordinates of standard button corners;
\item ${\textbf{M  =  [}}{{\textbf{m}}_1}{\rm{, \cdot \cdot \cdot , }}{{\textbf{m}}_N}{\rm{]  }} \in {R^{3 \times N}}$ - new spatial coordinates of detected button corners after rotation operation;
\item ${\textbf{P  =  [}}{{\textbf{p}}_1}{\rm{, \cdot \cdot \cdot , }}{{\textbf{p}}_N}{\rm{]  }} \in {R^{3 \times N}}$ - new spatial coordinates of detected button corners with depth equal to 1 after rotation and translation operation; 
\item ${\textbf{R}}\,({\rm{\theta }})$ - the matrix representation of angle-axis parameterized rotation ${\rm{\theta}}$;
\item ${\textbf{T}}$ - the matrix representation of translation, between detected button corners and standard button corners;
\item ${\rm{b}}$ - the number of buttons on image;
\item ${{\textbf{K}}_H}{\rm{  =  [}}{{\textbf{k}}_{h1}}{\rm{, \cdot \cdot \cdot , }}{{\textbf{k}}_{hN}}{\rm{]  }} \in {R^{1 \times N}}$ - slopes of the horizontal line of every button in space coordinate;
\item ${{\textbf{K}}_V}{\rm{  =  [}}{{\textbf{k}}_{v1}}{\rm{, \cdot \cdot \cdot , }}{{\textbf{k}}_{vN}}{\rm{]  }} \in {R^{1 \times N}}$ - slopes of the vertical line of every button in space coordinate;
\item ${\textbf{Cos  =  [Co}}{{\textbf{s}}_1}{\rm{, \cdot \cdot \cdot , }}{{\textbf{Cos}}_N}{\rm{]  }} \in {R^{1 \times N}}$ - cosine values of the angles between horizontal and vertical lines of every button in space coordinate.
\end{itemize}

The detail of the proposed perspective distortion removal algorithm is shown in Alg. \ref{pdr}.

\begin{algorithm}[t]
	\DontPrintSemicolon
	\SetKwRepeat{Do}{do}{while}
	\SetKwInOut{Input}{Input}
	\SetKwInOut{Output}{Output}
	\Input{$\textbf{C}, \textbf{U}, Distorted \; Image$}
	\Output{$Rectified \; Image$}
	$\mathop {{\textbf{C}}}\limits^ \wedge = Norm(\textbf{C}),$ $\mathop {{\textbf{U}}}\limits^ \wedge = Norm(\textbf{U});$ \\
	${\textbf{D}} = \,M_{{\mathop{\rm int}} }^{ - 1}\,{\rm{*}}\,\mathop {\textbf{C}}\limits^ \wedge,$
	${\textbf{E}} = \,M_{{\mathop{\rm int}} }^{ - 1}\,{\rm{*}}\,\mathop {\textbf{U}}\limits^ \wedge;$\\
	$Best \; CR = +\infty,$
	${\theta_x}^{'}={\theta_y}^{'}={\theta_z}^{'}=None;$\\
	\For{${\theta_x} \; in \; range(\alpha,\beta,\gamma)$}{
	\For{${\theta_y} \; in \; range(\alpha,\beta,\gamma)$}{
	\For{${\theta_z} \; in \; range(\alpha,\beta,\gamma)$}{
		${\rm{\textbf{R}(\theta }}) = \;{{\textbf{R}}_x}\,{\rm{*}}\,{{\textbf{R}}_y}\,{\rm{*}}\;{{\textbf{R}}_z};$\\
		${\textbf{M}} = \;{\textbf{R}}({\rm{\theta }})\;*\;{\textbf{D}};$\\
		${\textbf{P}}\; = \;{\textbf{M}}\; + \;{\textbf{T}},$
        ${\textbf{P}}\; = \;{\textbf{P}}\,{\textbf{/P}}\,{\rm{[3]}};$\\
        \If{$Final \; CR < Best \; CR$}{
                $Best \; CR = Final \; CR$\\
                ${\theta_x}^{'}={\theta_x},{\theta_y}^{'}={\theta_y},{\theta_z}^{'}={\theta_z};$\\
                ${{\textbf{R}}_x}^{'} \leftarrow {\theta_x}^{'},
                {{\textbf{R}}_y}^{'} \leftarrow {\theta_y}^{'},
                {{\textbf{R}}_z}^{'} \leftarrow {\theta_z}^{'};$\\
                ${\rm{\textbf{R}(\theta }})^{'} = \;{{\textbf{R}}_x}^{'}\,{\rm{*}}\,{{\textbf{R}}_y}^{'}\,{\rm{*}}\;{{\textbf{R}}_z}^{'};$\
	}
	}
	}
	}
	
	${\textbf{M}}^{'} = \;{\textbf{R}}({\rm{\theta }})^{'}\;*\;{\textbf{D}};$\\
	
	${\textbf{P}}^{'}\; = \;{\textbf{M}}^{'}\; + \;{\textbf{T}}^{'},$
	${\textbf{P}}^{'}\; = \;{\textbf{P}}^{'}\,{\textbf{/P}}^{'}\,{\rm{[3]}};$\\
	
	$\mathop {\textbf{G}}\limits^ \wedge   = {{\rm{M}}_{{\mathop{\rm int}} }}*{\textbf{P}}^{'};$
	
	$\text{Return} \;\mathop {\textbf{G}}\limits^ \wedge;$\\
			
	\caption{Perspective Distortion Removal}
	\label{pdr}
\end{algorithm}


The first step is to establish a presupposed elevator button image, in which pixel coordinates of the button corners \textbf{U} are standard without perspective distortion.
Two types of the presupposed elevator button images are shown in Fig. 4.
\begin{figure}[h]
\centering
\includegraphics[height=4.62cm,width=6.42cm]{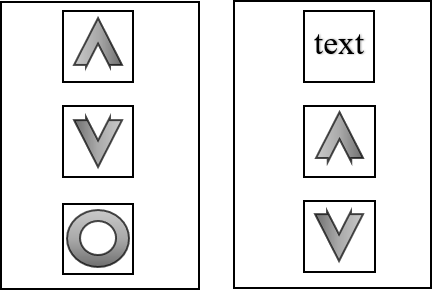}
\caption{Two types of the presupposed elevator button images without perspective distortion.}
\end{figure} 

The second step is back projection.
$\mathop {\textbf{C}}\limits^ \wedge  $ and $\mathop {\textbf{U}}\limits^ \wedge   \in {R^{{\rm{3}} \times N}}$ are obtained by adding a third row $\left[ {1 \cdot  \cdot  \cdot \;1} \right]$ to \textbf{C} and $\textbf{U} \in {R^{2 \times N}}$.
Then the inverse matrix of the intrinsic camera parameter is leveraged to obtain the spatial coordinates of button corners, and the equation is shown as follows:
\begin{equation}
    {\textbf{D}} = \,M_{{\mathop{\rm int}} }^{ - 1}\,{\rm{*}}\,\mathop {\textbf{C}}\limits^ \wedge, \label{eq}
\end{equation}
In this algorithm, we assume that for the standard button corners without perspective distortions, the slopes of horizontal lines equal to zero, the slopes of vertical lines equal to infinity, and the cosine values of the angles between horizontal and vertical lines equal to zero. Thus for \textbf{E}, we have:
\begin{equation}
    {{\textbf{K}}_{\rm{H}}}\, = \,1/\,{{\textbf{K}}_V}{\textbf{  =  Cos }} = {\rm{  [}}0{\rm{, \cdot \cdot \cdot , }}0{\rm{]  }} \in {R^{1 \times N}}. \label{eq}
\end{equation}

The third step is to compute the rotation and translation matrix to form new spatial coordinates of detected button corners. 
Three-dimensional rotation matrices are utilized to rotate spatial coordinates of the corners, which include rotating ${{\rm{\theta }}_x}$ against the x-axis, ${{\rm{\theta }}_y}$ against the y-axis and ${{\rm{\theta }}_z}$ against the z-axis, shown as follows respectively:
\begin{equation}
   {{\textbf{R}}_x}\, = \,\left[ \begin{array}{l}
\,\,1\quad \,\;\;\;\;0\quad \,\quad \quad \;\,0\\
\;0\quad \,\,\cos ({\theta _x})\quad \;\sin ({\theta _x})\\
\;0\quad  - \sin ({\theta _x})\;\;\;\;\,\cos ({\theta _x})
\end{array} \right], \label{eq}
\end{equation}

\begin{equation}
   {{\textbf{R}}_y}\, = \,\left[ \begin{array}{l}
\cos ({\theta _y})\quad \,0\quad \, - \sin ({\theta _y})\\
\;\;\,\,0\quad \,\,\;\quad \;1\quad \;\quad \;0\\
\sin ({\theta _y})\quad \,\,0\;\quad \;\cos ({\theta _y})
\end{array} \right], \label{eq}
\end{equation}

\begin{equation}
   {{\textbf{R}}_z}\, = \,\left[ \begin{array}{l}
\;\;\,\cos ({\theta _z})\quad \,\sin ({\theta _z})\quad \,0\\
 - \sin ({\theta _z})\quad \,\cos ({\theta _z})\quad \;0\\
\;\;\quad 0\quad \,\,\quad \quad \,0\;\quad \;\quad \;1
\end{array} \right], \label{eq}
\end{equation}
where ${{\rm{\theta }}_x}$, ${{\rm{\theta }}_y}$, ${{\rm{\theta }}_z}$  are radian values and the relation between angle value and radian value is:

\begin{equation}
   radian\;value\; = \;\frac{{angle\;value\;*\;\pi }}{{180}}. \label{eq}
\end{equation}
The rotation matrix is formed as ${\rm{\textbf{R}(\theta }}) = \;{{\textbf{R}}_x}\,{\rm{*}}\,{{\textbf{R}}_y}\,{\rm{*}}\;{{\textbf{R}}_z}.$
Then, new spatial coordinates of detected button corners are computed through the following equations:


\begin{equation}
    {\textbf{M}} = \;{\textbf{R}}({\rm{\theta }})\;*\;{\textbf{D}},
    {\textbf{P}}\; = \;{\textbf{M}}\; + \;{\textbf{T}}, {\textbf{P}}\; = \;{\textbf{P}}\,{\textbf{/P}}\,{\rm{[3]}},
    \label{eq7}
\end{equation}

where \textbf{P[3]} represents the third row of new spatial coordinates, and the translation matrix \textbf{T} is defined as the difference value between spatial coordinates of the first corner in the presupposed elevator button image and the distorted elevator button image.


The fourth step is to estimate camera motions.
The goal is to find the optimal rotation matrix and translation matrix so that the lines formed by the new spatial coordinates of the distorted button corners are parallel to the lines formed by the spatial coordinates of the presupposed standard button corners. In this algorithm, we set $\alpha=-40,\beta=40,\gamma=0.5$, which means that every 0.5 degree against each axis is sampled to form the rotation matrix ${\textbf{R}}({\rm{\theta }})$.
The range is from -40 degrees to 40 degrees.
Three criteria are chosen to evaluate which formed rotation matrix is optimal:

The first criterion is ${{\textbf{K}}_H}$, representing the slopes of horizontal lines of every button in space coordinate.
${{\textbf{k}}_{hi}}$ of each button is defined as:

\begin{equation}
    {{\textbf{k}}_{hi}} = \frac{{{y_2} - {y_1}}}{{{x_2} - {x_1}}},\label{eq}
\end{equation}
where ${x_i}$ denotes the first value of the $i$-th corner and ${y_i}$ denotes the second value of the $i$-th corner, respectively.
Then we can obtain the two-norm result of ${{\textbf{K}}_H}$,

\begin{equation}
    {\left\| {{{\textbf{K}}_H}} \right\|_2} = \sqrt {\sum\limits_{i = 1}^{\rm{b}} {{\textbf{k}}_{hi}^2} }.\label{eq}
\end{equation}

The second criterion is ${{\textbf{K}}_V}$, representing the slopes of vertical lines of every button in space coordinate. ${{\textbf{k}}_{vi}}$ of each button is defined as:
\begin{equation}
    {{\textbf{k}}_{vi}} = \frac{{{y_4} - {y_1}}}{{{x_4} - {x_1}}},\label{eq}
\end{equation}
Then we can obtain the two-norm result of ${{\rm{K}}_V}$ and its reciprocal ${{\textbf{K}}_{rV}}$,

\begin{equation}
    \begin{array}{l}
    {\left\| {{{\textbf{K}}_V}} \right\|_2} = \sqrt {\sum\limits_{i = 1}^{\rm{b}} {{\textbf{k}}_{vi}^2} },\\
    {{\textbf{K}}_{rV}} = 1\,{\rm{/}}\,{\left\| {{{\textbf{K}}_v}} \right\|_2}.
\end{array} \label{eq}
\end{equation}

The third criterion is \textbf{Cos}, representing the cosine values of the angles between horizontal and vertical lines of every button in space coordinate.
The horizontal line vector, vertical line vector, and ${\textbf{co}}{{\textbf{s}}_i}$ of each button are shown as follows:

\begin{equation}
    \begin{array}{l}
h = ({x_2} - {x_1},{y_2} - {y_1},{z_2} - {z_1}),\\
v = ({x_4} - {x_1},{y_4} - {y_1},{z_4} - {z_1}),\\
{\textbf{co}}{{\textbf{s}}_i}\; = \frac{{h \bullet v}}{{{{\left\| h \right\|}_2} \bullet {{\left\| v \right\|}_2}}},
\end{array}\label{eq}
\end{equation}

\begin{figure*}
    \centering
    \subfigure[]{
    \begin{minipage}[t]{0.18\linewidth}
    \centering
    \includegraphics[height=3.4cm,width=3cm]{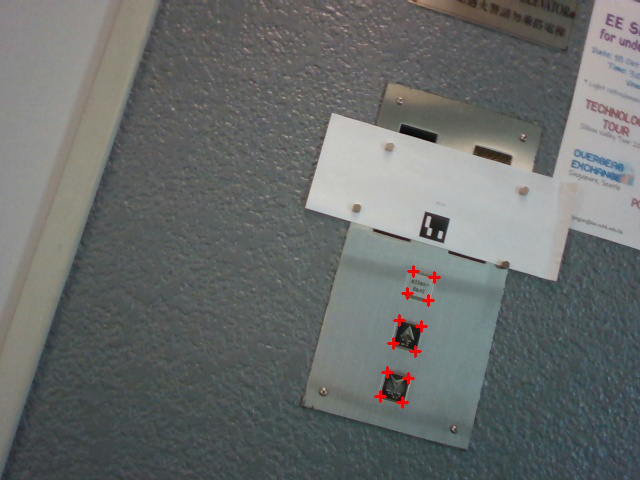}
    \end{minipage}}
    \subfigure[]{
    \begin{minipage}[t]{0.18\linewidth}
    \centering
    \includegraphics[height=3.4cm,width=3cm]{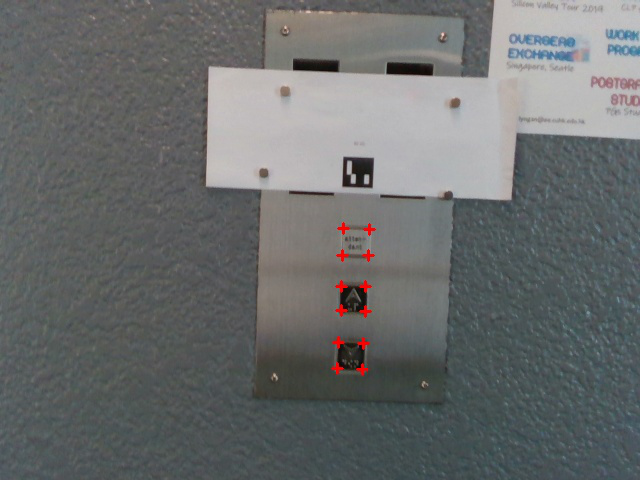}
    \end{minipage}}
    \subfigure[]{
    \begin{minipage}[t]{0.18\linewidth}
    \centering
    \includegraphics[height=3.4cm,width=3cm]{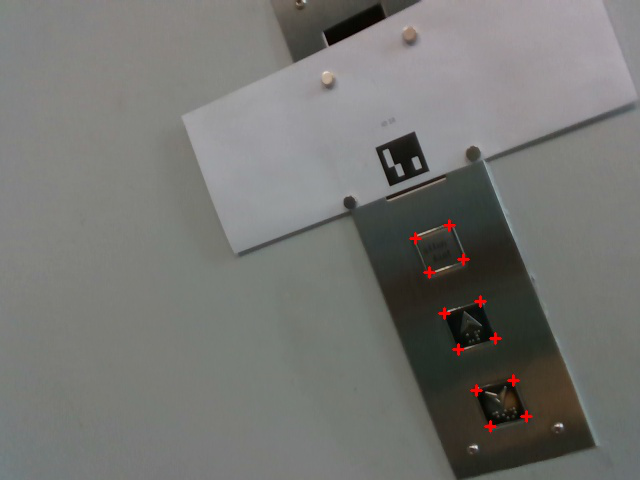}
    \end{minipage}}
    \subfigure[]{
    \begin{minipage}[t]{0.18\linewidth}
    \centering
    \includegraphics[height=3.4cm,width=3cm]{./imgs/figure5d.png}
    \end{minipage}}
    \subfigure[]{
    \begin{minipage}[t]{0.18\linewidth}
    \centering
    \includegraphics[height=3.4cm,width=3cm]{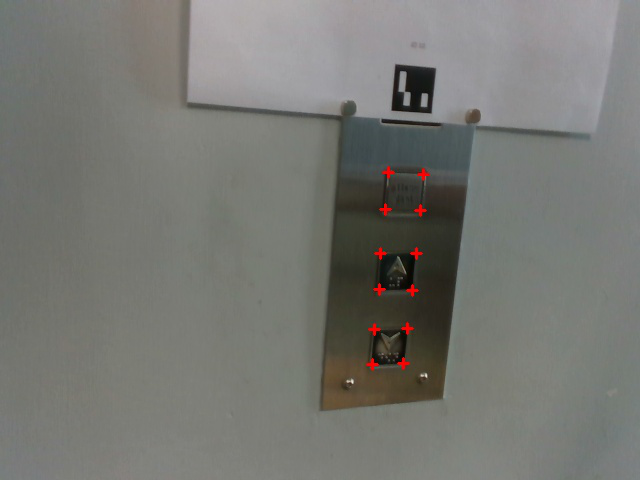}
    \end{minipage}}
    \\\subfigure[]{
    \begin{minipage}[t]{0.18\linewidth}
    \centering
    \includegraphics[height=3.4cm,width=3cm]{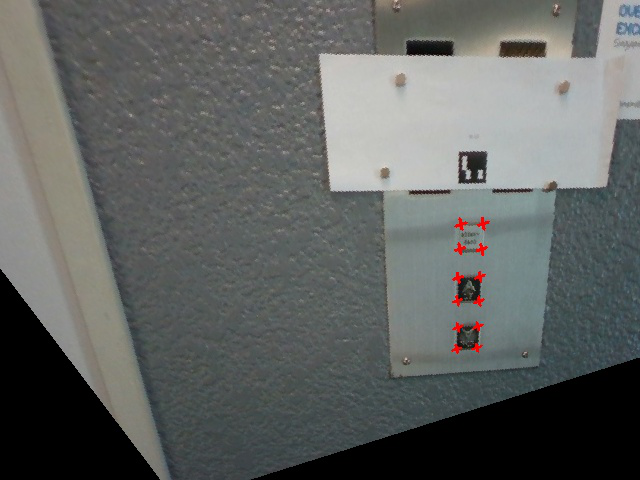}
    \end{minipage}}
    \subfigure[]{
    \begin{minipage}[t]{0.18\linewidth}
    \centering
    \includegraphics[height=3.4cm,width=3cm]{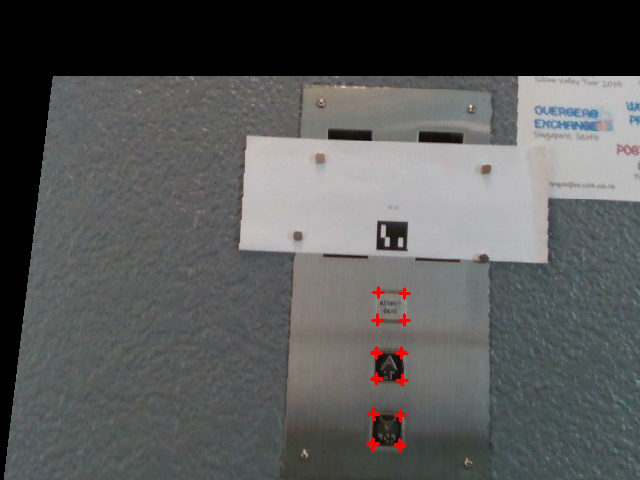}
    \end{minipage}}
    \subfigure[]{
    \begin{minipage}[t]{0.18\linewidth}
    \centering
    \includegraphics[height=3.4cm,width=3cm]{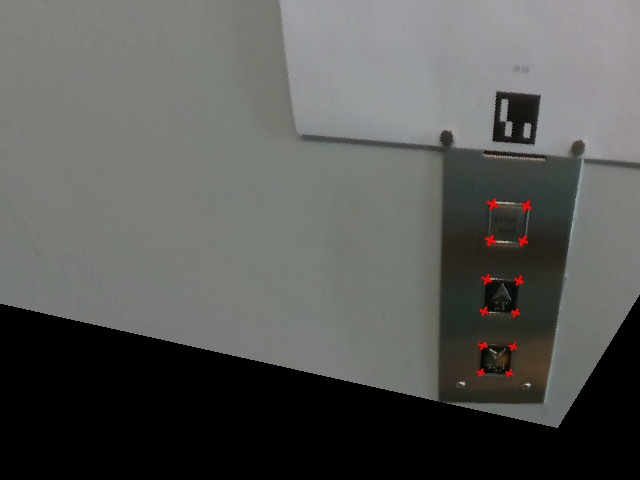}
    \end{minipage}}
     \subfigure[]{
    \begin{minipage}[t]{0.18\linewidth}
    \centering
    \includegraphics[height=3.4cm,width=3cm]{./imgs/figure5i.png}
    \end{minipage}}
    \subfigure[]{
    \begin{minipage}[t]{0.18\linewidth}
    \centering
    \includegraphics[height=3.4cm,width=3cm]{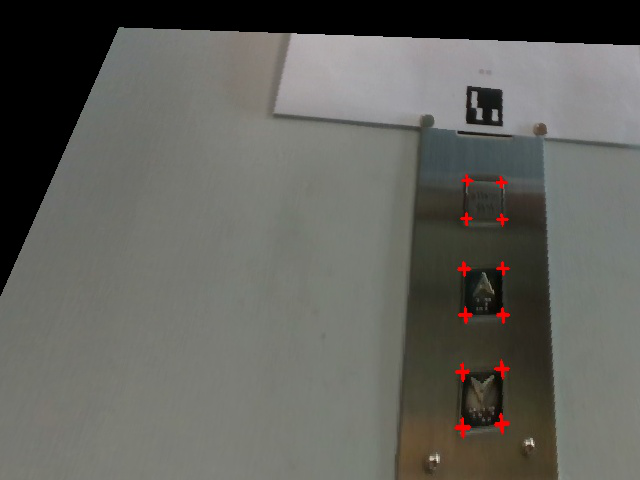}
    \end{minipage}}
    \centering
    \caption{Demonstrations of perspective distortion removal results. (a)(b)(c)(d)(e) 
    represent the original elevator button images, (f)(g)(h)(i)(j) 
represent the rectified elevator button images, respectively.}
    \label{fig:important}
\end{figure*}

where $({x_1},{y_1},{z_1})$ denotes the spatial coordinate of the first corner, $({x_2},{y_2},{z_2})$ the spatial coordinate of the second corner, and $({x_4},{y_4},{z_4})$ the spatial coordinate of the fourth corner. Then we can obtain the two-norm result of \textbf{Cos},
\begin{equation}
    {\left\| {{\textbf{Cos}}} \right\|_2} = \sqrt {\sum\limits_{i = 1}^{\rm{b}} {{\textbf{cos}}_i^{\rm{2}}} }. \label{eq13}
\end{equation}
\\When ${\left\| {{\textbf{Cos}}} \right\|_2}$ is smaller, we can obtain a better perpendicular result of horizontal and vertical lines of buttons.

To combine the three criteria, they are normalized as follows:

\begin{equation}
    {\mathop {\textbf{K}}\limits^ \wedge  _H} = \frac{{{{\left\| {{{\textbf{K}}_H}} \right\|}_2} - {{\left( {{{\left\| {{{\textbf{K}}_H}} \right\|}_2}} \right)}_{\min }}}}{{{{\left( {{{\left\| {{{\textbf{K}}_H}} \right\|}_2}} \right)}_{\max }} - {{\left( {{{\left\| {{{\textbf{K}}_H}} \right\|}_2}} \right)}_{\min }}}},\label{eq}
\end{equation}

\begin{equation}
    {\mathop {\textbf{K}}\limits^ \wedge  _{rV}} = \frac{{{{\left\| {{{\textbf{K}}_{rV}}} \right\|}_2} - {{\left( {{{\left\| {{{\textbf{K}}_{rV}}} \right\|}_2}} \right)}_{\min }}}}{{{{\left( {{{\left\| {{{\textbf{K}}_{rV}}} \right\|}_2}} \right)}_{\max }} - {{\left( {{{\left\| {{{\textbf{K}}_{rV}}} \right\|}_2}} \right)}_{\min }}}},\label{eq}
\end{equation}

\begin{equation}
    \mathop {{\textbf{Cos}}}\limits^ \wedge   = \frac{{{{\left\| {{\textbf{Cos}}} \right\|}_2} - {{\left( {{{\left\| {{\textbf{Cos}}} \right\|}_2}} \right)}_{\min }}}}{{{{\left( {{{\left\| {{\textbf{Cos}}} \right\|}_2}} \right)}_{\max }} - {{\left( {{{\left\| {{\textbf{Cos}}} \right\|}_2}} \right)}_{\min }}}}.\label{eq}
\end{equation}
Then the final criterion is shown as follows:
\begin{equation}
    Final\;CR = {\mathop {\textbf{K}}\limits^ \wedge  _H} + {\mathop {\textbf{K}}\limits^ \wedge  _{rV}} + \mathop {{\textbf{Cos}}}\limits^ \wedge. \label{eq}
\end{equation}
When $Final CR$ is smallest, we can obtain the optimal rotation matrix and translation matrix.

The fifth step is to form new rectified images. After obtaining the optimal pose (${\textbf{R}}({\rm{\theta }})^{'},{\textbf{T}}^{'}$), each pixel of the distorted elevator button image can be transformed to have new spatial coordinates through Eq. (\ref{eq7}). 
Then intrinsic camera parameter is used to do projection and get new pixel coordinates in the normalized plane, 
\begin{equation}
    \mathop {\textbf{G}}\limits^ \wedge   = {{\rm{M}}_{{\mathop{\rm int}} }}*{\textbf{P}}.\label{eq}
\end{equation}
By taking the first and second rows, we can obtain pixel coordinates of the rectified button corners in the image plane.
Finally, a new rectified elevator button image is generated by applying an inverse image warping operation.

\section{Experiments}
To verify the effectiveness of the proposed approach, we collect a dataset with 15 images from 3 different elevators, which are captured from different angles of views containing severe perspective distortions. The intrinsic camera parameter is:

\begin{equation}
    {{\rm{M}}_{{\mathop{\rm int}} }}^{'}\, = \,\left[ \begin{array}{l}
    \;{\rm{320}}\quad \,{\rm{0}}\quad \,\;\;{\rm{320}}\\
    \;\;\;{\rm{0}}\quad \,{\rm{320}}\quad \,{\rm{240}}\\
    \;\;\;{\rm{0}}\quad \,\;\;{\rm{0}}\quad \,\;\;\;\;\;\,{\rm{1}}
\end{array} \right]
\end{equation}
The value of Eq. (\ref{eq13}) is used to measure the accuracy of the proposed perspective distortion removal algorithm, which represents the two-norm value of cosine values of the angles between horizontal and vertical lines of all buttons in space coordinate. When the value of Eq. (\ref{eq13}) is smaller, the rectification performance is better.
The experimental results of 15 elevator button images are shown in Table \ref{table}.
Some demonstrations of the corresponding original and rectified images are presented in Fig. \ref{fig:important}. 
From Fig. \ref{fig:important} and Table \ref{table}, we can see that our proposed approach is capable of removing perspective distortions of elevator button images autonomously with high accuracy.

\begin{table}[htbp]
  \centering
  \caption{Accuracy of distortion removal}
    \begin{tabular}{ccccccc}
    \toprule
    No.   & I-10  & I-20  & I-30  & I-40  & I-50  & Average \\
    1     & 0.036  & 0.042  & 0.050  & 0.007  & 0.024  & 0.032  \\
    \midrule
    No.   & I-160 & I-170 & I-180 & I-190 & I-200 & Average \\
    2     & 0.003  & 0.026  & 0.003  & 0.004  & 0.016  & 0.010  \\
    \midrule
    No.   & I-850 & I-860 & I-870 & I-880 & I-890 & Average \\
    3     & 0.048  & 0.070  & 0.097  & 0.100  & 0.074  & 0.078  \\
    \bottomrule
    \end{tabular}%
  \label{table}%
\end{table}%

\section{Conclusion}

This article proposes a novel deep learning-based approach that can autonomously remove perspective distortions of elevator button images.
We utilize an image segmentation model and Hough Transform method to obtain the detection results of button corners.
A novel algorithm is designed to correct the perspective distortions of original elevator button images.
Currently, the presented algorithm can only handle elevator button images that contain several rectangle buttons. 
For elevator button images that contain circular buttons, as a circle has no slopes, the presented algorithm will fail. 
In the next step, we will use the Hough transform method to calculate the center coordinates and radius of the circular buttons and develop a novel algorithm that can autonomously remove perspective distortions of images containing circular elevator buttons.

%
%
\bibliographystyle{IEEEtran}
\bibliography{nc.bib} 

\end{document}